\documentclass[11pt,a4paper]{article}
\usepackage[hyperref]{eacl2021}
\usepackage{times}
\usepackage{latexsym}
\usepackage{enumitem}

\usepackage{microtype}

\aclfinalcopy 
\def\aclpaperid{47} 

\title{Towards Personalised and Document-level \\ Machine Translation of Dialogue}

\author{Sebastian T. Vincent \\
  Department of Computer Science, University of Sheffield \\
  Regent Court, 211 Portobello, Sheffield, S1 4DP, UK \\
  \texttt{stvincent1@sheffield.ac.uk} }

\date{}

\begin{document}
\maketitle
\begin{abstract}
State-of-the-art (SOTA) neural machine translation (NMT) systems translate texts at sentence level, ignoring context: intra-textual information, like the previous sentence, and extra-textual information, like the gender of the speaker. Because of that, some sentences are translated incorrectly. Personalised NMT (\mbox{PersNMT}) and document-level NMT (\mbox{DocNMT}) incorporate this information into the translation process. Both fields are relatively new and previous work within them is limited. Moreover, there are no readily available robust evaluation metrics for them, which makes it difficult to develop better systems, as well as track global progress and compare different methods. This thesis proposal focuses on PersNMT and DocNMT for the domain of dialogue extracted from TV subtitles in five languages: English, Brazilian Portuguese, German, French and Polish. Three main challenges are addressed: (1) incorporating extra-textual information directly into NMT systems; (2) improving the machine translation of cohesion devices; (3) 
reliable evaluation for PersNMT and DocNMT. 
\end{abstract}

\section{Introduction}
Neural machine translation (NMT) represents state-of-the-art (SOTA) results in many domains \citep{Sutskever2014, Vaswani2017, Lample2020}, with some authors claiming human parity \citep{Hassan2018}. However, traditional methods process texts in short units like the utterance or sentence, isolating them from the entire dialogue or document, as well as ignoring extra-textual information (e.g. who is speaking, who they are talking to). This can result in a translation hypothesis' meaning or function being significantly different from the reference or make the text incohesive or illogical.
For instance, the sentence in Polish \textit{\enquote{Nie poszłam.}} (\textit{\enquote{I didn't go.}}\footnote{All examples throughout the report have been generated using Google Translate \url{http://translate.google.com/}, accessed 26 Nov 2020.}) incorporates gender information in the word \textit{poszłam} (\textit{went\textsubscript{fem}}) -- as opposed to \textit{poszedłem} (\textit{went\textsubscript{masc}}) -- while the English verb does not incorporate such information. When translating \textit{\enquote{I didn't go.}} into Polish, the machine translation (MT) model must guess the gender of \textit{I}, as this information is not rendered in the English sentence.  \citet{rescigno-etal-2020-case} show that when commercial MT engines need to \enquote{guess} the gender of a word, they do so by making implications based on its co-occurrence with other words in the training data. Since training data is often biased \citep{Stanovsky2020Bias}, MT models will reproduce these biases, further propagating and reinforcing them. Clearly, research on context-aware machine translation is needed.

Sentence-level NMT (SentNMT) is especially harmful in the domain of dialogue, where most utterances rely on previously spoken ones, both in content and in style. The way in which an interlocutor chooses to express themselves depends on what they perceive as the easiest for the other person to understand \citep{Pickering2004}. Dialogue is naturally cohesive \citep{Halliday2013}, i.e. rid of redundancies, confusing redefinition of terms and unclear references. Part of what makes a conversation fluent is the links between its elements, which SOTA NMT models fail to capture. For instance, the latter utterance in the following exchange: \textit{\enquote{They put something on the roof.} \enquote{What?}} translates to Polish as \textit{\enquote{Co takiego?} (\enquote{What something?})}. The translation uses information unavailable in the utterance itself, i.e. the fact that \textit{what} refers to the noun \textit{something}. A sentence-level translation of \textit{What?} would just be \textit{Co?}, which is more universal, but also more ambiguous. Simply put, even when SentNMT produces a feasible translation, its context agnosticism may prevent it from producing a far better one.

There are growing appeals for developing NMT systems capable of incorporating additional information into hypothesis production: personalised NMT for extra-textual information \citep[e.g.][]{Sennrich2016a, Elaraby2018, Vanmassenhove2020} and document-level NMT for intra-textual information \citep[e.g.][]{Bawden2019, tiedemann-scherrer-2017-neural, Zhang2020a, lopes-etal-2020-document}. Evaluation methods predominant within both areas vary vastly from paper to paper, suggesting that for these applications a robust evaluation metric is not readily available. This view is further strengthened by the fact that \citet{Hassan2018}, when assessing their MT for human parity, ignored document-level evaluation completely. \citet{Laubli2018} later disputed this choice, showing that professional annotators still overwhelmingly prefer human translation at the level of the document, and therefore human parity has not yet been achieved. This case study shows how much a robust and widely accepted document-level metric is needed.

Currently, researchers working on PersNMT and DocNMT conduct evaluation primarily by reporting the BLEU score for their systems. But they also commonly assert that the metric cannot reliably judge fine-grained translation improvements coming from context inclusion. As a way out, some of them report accuracy on specialised test suites \citep[e.g.][]{kuang-etal-2018-modeling, Bawden2019, Voita2020} or manual evaluation. Although both have limited potential for generalisation, their attention to detail makes them superior tactics of evaluation for applications such as PersNMT and DocNMT.

In this work we utilise TV subtitles, a context-rich domain, in order to investigate whether MT of dialogue can be improved: directly, by \textbf{enhancing document coherence and cohesion} through incorporation of intra- and extra-textual information into translation, and indirectly, by \textbf{designing suitable evaluation methods for PersNMT and DocNMT}. Dialogue extracted from TV content is an attractive domain for two reasons: (1) there is an abundance of parallel dialogue corpora extracted purely from subtitles, and (2) the data is rich in or could potentially be annotated for a range of meta information such as the gender of the speaker. 

In Section 2, we discuss relevant contextual phenomena. We then present the research on PersNMT and DocNMT, and the applicability of MT evaluation metrics to both. In Section 3 we delineate the research questions, the work conducted so far and our plans. Section 4 concludes the paper.

\section{Background}
\subsection{Contextual phenomena}
Two types of contextual phenomena relevant for MT of dialogue are explored: \textbf{cohesion} phenomena (related to information that can be found in the text) and \textbf{coherence} phenomena (related to the context of situation, which we consider to be external to the text). We emphasise that the phenomena explored below represent a subset of cohesion and coherence constituents, and that our interest in them arises from the difficulties they pose for MT of dialogue.

\paragraph{Cohesion phenomena} Humans introduce cohesion into speech or written text in three ways: by choosing words related to those that were used before (\textit{lexical cohesion}), by omitting parts of or whole phrases which can be unambiguously recovered by the addressee (\textit{ellipsis and substitution}) and by referring to elements with pronouns or synonyms that the speaker judges recoverable from somewhere else in text (\textit{reference}) \citep{Halliday2013}. 
Cohesion phenomena effectively constitute links in text, whether within one utterance or across several. \autoref{tab:cohesion_ex} shows examples of how they can be violated by MT.

\begin{figure}[ht]
\centering
\begin{tabular}{p{0.1\linewidth}p{0.80\linewidth}}
\textbf{EN}  & \enquote{It's just a \underline{social call}.} \enquote{A \underline{social call}?}            \\
        \textbf{PL\textsubscript{MT}} & \enquote{To tylko \textcolor{BrickRed}{spotkanie towarzyskie}.}
        \enquote{\textcolor{BrickRed}{Połączenie towarzyskie}?} \\
        & \textit{(\enquote{It's just a social gathering.} \enquote{A social call?})}\\
        \textbf{PL\textsubscript{ref}}  & \enquote{To tylko \textcolor{OliveGreen}{spotkanie towarzyskie}.} \enquote{\textcolor{OliveGreen}{Spotkanie towarzyskie}?}   \\
        & \textit{(\enquote{It's just a social gathering.} \enquote{A social gathering?})}\\ \hline
\textbf{EN}  & I \underline{love it}. We all \underline{do} [=love it].                              \\
\textbf{PL\textsubscript{MT}} & \textcolor{BrickRed}{Kocham} to. Wszyscy \textcolor{BrickRed}{to robimy}. \textit{(\enquote{We all do it.})}                \\
\textbf{PL\textsubscript{ref}}  & \textcolor{OliveGreen}{Kocham} to. Wszyscy \textcolor{OliveGreen}{to kochamy}. \textit{(\enquote{We all love it.})}
\end{tabular}
\caption{Mistranslations of cohesion phenomena in translations. In the top example, \textit{social call} is reiterated in source and reference, while MT opts for two different phrases, thereby decreasing lexical cohesion. The bottom example is verb phrase ellipsis, which does not exist in Polish and hence requires that the antecedent verb is repeated.}
\label{tab:cohesion_ex}
\end{figure}

Cohesion-related tasks such as coreference or ellipsis resolution have attracted great interest in the recent years \citep[e.g.][]{Rnning2018, Jwalapuram2020}. Previous research on cohesion within DocNMT has revealed that verb phrase ellipsis, coreference and reiteration (a type of lexical cohesion) may be particularly erroneous in MT \cite[e.g.][]{tiedemann-scherrer-2017-neural, Bawden2018, Voita2020}. 

\paragraph{Coherence phenomena} Coherence is consistency of text with the context of situation \cite{Halliday76a}. MT of dialogue may be erroneous due to models not having access to extra-textual information\footnote{Note: the focus here is on \textit{sentence-level translation utilising extra-textual context.}}, e.g.: (a) speaker gender and number, (b) interlocutor gender and number, (c) social addressing, and (d) discourse situation. Different languages may render such phenomena differently, e.g. formality in German is expressed through the formal pronoun \textit{Sie} (e.g. \enquote{Are you hungry?} becomes \enquote{\textit{Bist du hungrig?}} when informal and \enquote{\textit{Sind Sie hunrgig?}} when formal), while in Polish inflections of the pronoun \textit{Pan/Pani/Państwo} (\enquote{\textit{Mr/Mrs/Mr and Mrs}}), the formal equivalent of \textit{ty/wy} (\enquote{\textit{you}}) are used. Then, as observed by \citet{Kranich2014}, some languages (such as English) prefer to express formality through politeness via word choices (e.g. \textit{pleased} is a more formal \textit{happy})\footnote{More examples can be found in the Appendix}. 

\subsection{Personalised Neural Machine Translation}
\label{rev:persnmt}

In PersNMT, the aim is to develop a system $F$ capable of executing the following operation:
\begin{equation*}
    F(x_{SL}, e, TL) = x_{TL, e}
\end{equation*}
where $x$ is the source sentence, $p$ is the extra-textual information (e.g. speaker gender) and $SL, TL$ are source and target language, respectively; $x_{TL, e}$ is then a contextual translation of $x_{SL}$.

This formulation is inspired by previous work within the area. \citet{Sennrich2016a} control the formality of a sentence translated from English to German by using a side constraint. The model is trained on pairs of sentences $(x_i, y_i)$, where $y_i$ is either formal or informal, and a corresponding tag is prepended to the source sentence. At test time, the model relies on the tag to guide the formality of the translation hypothesis. A similar method has been used in \citet{Vanmassenhove2020} and in \citet{Elaraby2018} to address the problem of speaker gender morphological agreement. \citet{moryossef-etal-2019-filling} address the issue by modifying the source sentence during inference. They prepend the source with a minimal phrase implicitly containing all the relevant information; for example, for a female speaker and a plural audience, the augmented source yields \enquote{\textit{She said to them: <src. sent.>}}. Their method improves on multiple phenomena simultaneously (speaker gender and number, interlocutor gender and number) and requires little annotated data, but its performance relies entirely on the MT system's ability to utilise the added information. Furthermore, there are some side effects, e.g. the authors find the model's predictions to be often unintentionally influenced by the token \textit{said}.

A similar method of tag-managed tuning has been used to train multilingual NMT systems \citep{Johnson2017} and approximately control sequence length in NMT \citep{Lakew2019}. Outside MT, this method has been the driving force behind large pretrained controllable language models \citep{devlin-etal-2019-bert, Keskar2019ctrl, Dathathri2019, Krause2020, Mai2020pplm}.

\subsection{Document-level Neural Machine Translation (DocNMT)}
Traditionally, NMT is a sentence-level (\textbf{Sent2Sent}) task, where models process each sentence of a document independently. Another way to do it would be to process the entire document at once (\textbf{Doc2Doc}), but it is much harder to train a reliable NMT model on document-long sequences. A compromise between the two is a \textbf{Doc2Sent} approach which produces the translation sentence by sentence but considers the document-level information as \textit{context} when doing so \citep{Sun2020}.

\paragraph{Doc2Doc} \citet{tiedemann-scherrer-2017-neural} conduct the first Doc2Doc pilot study: they translate documents two sentences at once, each time discarding the first translated sentence and keeping the latter. They find that there is some benefit from doing so, albeit such benefit is difficult to measure. A larger setting was explored in \citep{junczys-dowmunt-2019-microsoft}: a 12-layer Transformer-Big \citep{Vaswani2017} was trained to translate documents of up to 1000 subword units, with performance optimised by noisy back-translation, fine tuning and second-pass post editing described in  \citep{junczys-dowmunt-grundkiewicz-2018-ms}. Finally, \citet{Sun2020} propose a fully Doc2Doc approach applicable to documents of arbitrary length. They split each document into $k \in {1,2,4,8...}$ parts and treat them as input data to the model, in what they call a \textbf{multi-relational} training, as opposed to \textbf{single relational} where only the whole document would be fed as input. Despite good results, the last two methods require enormous computational resources, and this limits their commercial application.

\paragraph{Doc2Sent} When translating a sentence $s_i$ a Doc2Sent model is granted access to document-level information $S \subseteq \{s_{0} ... s_{i-1}, s_{i+1} ... s_{n}\}$ and/or $T \subseteq \{t_{0} ... t_{i-1}\}$ where $n$ is the length of the document. The context information is either concatenated with the source sentence yielding a \textit{uni-encoder} model \citep{tiedemann-scherrer-2017-neural, ma-etal-2020-simple}, or is supplied in an extra encoder yielding a \textit{dual-encoder}\footnote{Notation adopted from \citet{ma-etal-2020-simple}.} model \citep{Zhang2020a, Voita2020}. In most approaches, the performance is optimised when shorter context (1-3 sentences) is used, though \citet{kim-etal-2019-document} find that applying a simple rule-based \textit{context filter} can stabilise performance for longer contexts. \citet{ma-etal-2020-simple} offer an improvement to uni-decoder which limits the sequence length in the top blocks of the Transformer encoder in the uni-encoder architecture, and \citet{kang-etal-2020-dynamic} introduce a reinforcement-learning-based \textit{context scorer} which dynamically selects the context best suited for translating the critical sentence.

\citet{jauregi-unanue-etal-2020-leveraging} challenge the idea that DocNMT can implicitly learn document-level features, and instead propose that the models be \textit{rewarded} when it preserves them. They focus on lexical cohesion and coherence and use respective metrics \citep{Wong2012, Gong2015} to measure rewards. This method may be successful provided that suitable specialised evaluation metrics are proposed in the future. Nevertheless, more interest has been expressed in literature in achieving high performance w.r.t. such features as a by-product of an efficient architecture, as is the case with SOTA Sent2Sent architectures.

\paragraph{Other architectures} DocRepair \citep{voita-etal-2019-context} is a monolingual post-editing model trained to repair cohesion in a document translated with SentNMT. \citet{kuang-etal-2018-modeling} use two cache structures to influence the model's token predictions: a dynamic cache $c_d$ of past token hypotheses with stopword removal and a topic cache $c_t$ of most probable topic-related words. Finally, \citet{lopes-etal-2020-document} compress the entire document into a vector and supply it as context during translation.

\subsection{Evaluation of Machine Translation}

Many machine translation evaluation (MTE) metrics have been proposed over the years, much owing to the yearly WMT Metrics task \citep{mathur-EtAl:2020:WMT}. They typically measure similarity between reference $r$, hypothesis $h$ and source $s$, expressed in e.g. n-gram overlap \citep[e.g.][]{papineni-etal-2002-bleu}, cosine distance of embeddings \citep[e.g.][]{DBLP:conf/iclr/ZhangKWWA20}, translation edit rate \citep{Snover2006} or trained on human judgements \citep{shimanaka-etal-2018-ruse}, with the SOTA represented by COMET which combines the ideas of \citeauthor{DBLP:conf/iclr/ZhangKWWA20} and \citeauthor{shimanaka-etal-2018-ruse}: several distances between $h, r$ and $s$ are computed based on contextual embeddings from BERT. 

Practically all of these metrics are developed to optimise performance at sentence level, an issue which until recently was not brought up often enough within the community. In the latest edition of the Metrics task at WMT \citep{mathur-EtAl:2020:WMT}, a track for document-level evaluation was introduced. However, the organisers approached document-level evaluation as the average of human judgements on sentences in documents. 
This is not a reliable assessment, since the quality of a text is more than the sum or average of the quality of its sentences. This approach risks \enquote{averaging out} the severity of potential inter-sentential errors. 
Currently, DocNMT models are typically evaluated in terms of BLEU, showing modest improvements over a baseline \citep[e.g.][report 0.7 BLEU improvement]{voita-etal-2018-context}. Several authors have argued that BLEU is not well suited to evaluating performance with respect to preserving cross-sentential discourse phenomena \citep{Voita2020, lopes-etal-2020-document}. When applied to methods which improve only a certain aspect of translation, BLEU can indicate very little about the accuracy of these improvements. Furthermore, \citet{kim-etal-2019-document} and \citet{li-etal-2020-multi-encoder} argue that even the reported BLEU gains in DocNMT models may not come from document-level quality improvements. \citet{li-etal-2020-multi-encoder} show that feeding the incorrect context can improve the metric by a similar amount. 

To decide whether DocNMT yield any improvements, a more sophisticated evaluation method is needed. Following the observation that DocNMT improves on individual aspects of translation w.r.t. SentNMT, \textbf{test suites} grew in popularity among researchers \citep{Bawden2019, Voita2020, lopes-etal-2020-document}. In particular, contrastive test suites \citep{muller-etal-2018-large} measure whether a model can repeatedly identify and correctly translate a certain phenomenon. They can be seen as robust collections of fine-grained multiple choice questions, yielding for each phenomenon an accuracy score indicative of performance. Producing these suites is time consuming and often requires expertise, but they are of extreme benefit to NMT. A sufficiently rich bed of test suites can evaluate the general robustness of a model, expressed as the average accuracy on these suites.

\section{Addressing Research Questions}
Within this PhD, we seek to answer three research questions (RQs):
\begin{enumerate}[noitemsep] 
    \item [RQ1] Can machine translation of dialogue be personalised by supplying it with extra-textual information?
    \item [RQ2] Is ellipsis problematic for MT, and can MT make use of marking of ellipsis and other cohesion devices to increase cohesion in translation of dialogue?
    \item [RQ3] How can automatic evaluation methods of MT be developed which confidently and reliably reward successful translations of contextual phenomena and, likewise, punish incorrect translations of the same phenomena?

\end{enumerate}

\subsection{Modelling Extra-Textual Information in Machine Translation}

We hypothesise that supplying the MT model with extra-textual information might help it make better dialogue translation choices. Our hypothesis is motivated by two facts: (1) that human translators base their choices of individual utterances on the understanding of the discourse situation and ensure that each utterance preserves its original function and meaning, and (2) that many instances of utterances and phrases are impossible to interpret unambiguously in isolation from their context. 

\paragraph{Tuning MT output with external information} 

Previous works on supplying context via constraints or tags have been narrow in scope, predominantly employing tag controlling (see \autoref{rev:persnmt}). Following their success we plan to experiment with alternative neural model architectures which allow the incorporation of extra data into sequence-to-sequence transduction and assess whether they are fit for translation. If successful, we see many potential applications of such models in NMT, ranging from those explored in this thesis to limiting the length of the translation, fine-grained personalisation (e.g. on speaker characteristics) and more.

\paragraph{Per scene domain adaptation} Neural machine translation models can be fine-tuned to a particular domain (e.g. medical transcripts) via domain adaptation \citep{Cuong2017Adaptation}. Effective as it is, domain adaptation requires domain-specific data and that the model is trained on it (a time-consuming process). This technique is then inapplicable in scenarios where domains are fine-grained and the adaptation needs to be instantaneous. Per scene adaptation appears to be a promising solution to the problem of wrong lexical choices made by MT models when translating dialogue. The environment or scene in which dialogue occurs is often crucial to interpreting its meaning; a scene-unaware model may misinterpret the function of an utterance and produce an incorrect translation.

Within TV dialogue we define a scene as continuous action which sets boundaries for exchanges. Its characteristics can be expressed in natural language (e.g. extracts from plot synopsis), as tags (e.g. \textit{school, student, sunny, exam}) or as individual categories (e.g. \textit{battle}). Since scene context is document-level, this task can also be seen as a use case for combining PersNMT and DocNMT, and will be explored in this PhD.


\begin{figure*}[t]
\centering
\begin{tabular}{ll}
\textbf{EN}  & \enquote{I'm sorry, Dad, but you \underline{wouldn't understand}.} \enquote{Oh, sure, I \underline{would [understand]}, princess.}                                                                                                     \\
\textbf{PL\textsubscript{MT}} & \enquote{Przepraszam tato, ale \textcolor{BrickRed}{nie zrozumiałbyś}.} \enquote{Och, \textcolor{BrickRed}{oczywiście}, księżniczko.}                                                                                                                      \\
\textbf{PL\textsubscript{ref}}  & \enquote{Przykro mi, tato, ale \textcolor{OliveGreen}{nie zrozumiałbyś}.} \enquote{\textcolor{OliveGreen}{Pewnie, że zrozumiałbym}, księżniczko.}                                                                                                       
\end{tabular}
\caption{A wrongly translated exchange with ellipsis. In the source, the word \textit{would} is a negation to \textit{wouldn't} in the previous utterance. The MT system ignores \textit{I would}: the backtranslation of PL\textsubscript{MT} reads \textit{\enquote{Oh, sure, princess.}}}
\label{tab:ellipsis_ex}
\end{figure*}


\subsection{Improving Cohesion for Machine Translation of Dialogue}
Work within MT so far has only limitedly explored whether ellipsis poses a significant problem for translation \citep[see][]{Voita2020}. We hypothesise that this is indeed the case: for some language pairs, the quality of machine-translated texts depend on the system's understanding of the ellipsis, when it is present in the source text. Since in dialogue ellipsis typically spans more than one utterance, it is poorly understood by SentNMT and the resulting MT quality is low (\autoref{tab:ellipsis_ex}).

To test our hypothesis, we will analyse ellipsis occurrences in dialogue data. We will use automatic methods to identify 1,000 occurrences of ellipsis in source text and mark spans of their occurrence in the corresponding machine and reference translations. All cases will then be manually analysed from the following angles: (i) Is the ellipsis correctly translated? (ii) Is the resulting translation of ellipsis natural/unnatural? (iii) Does the reference translation make use of the elided content? (iv) If the model generates an acceptable translation, could the elided content nevertheless have been used to disambiguate it or make it more cohesive? 

\begin{figure}[ht]
\centering
\includegraphics[width=\linewidth]{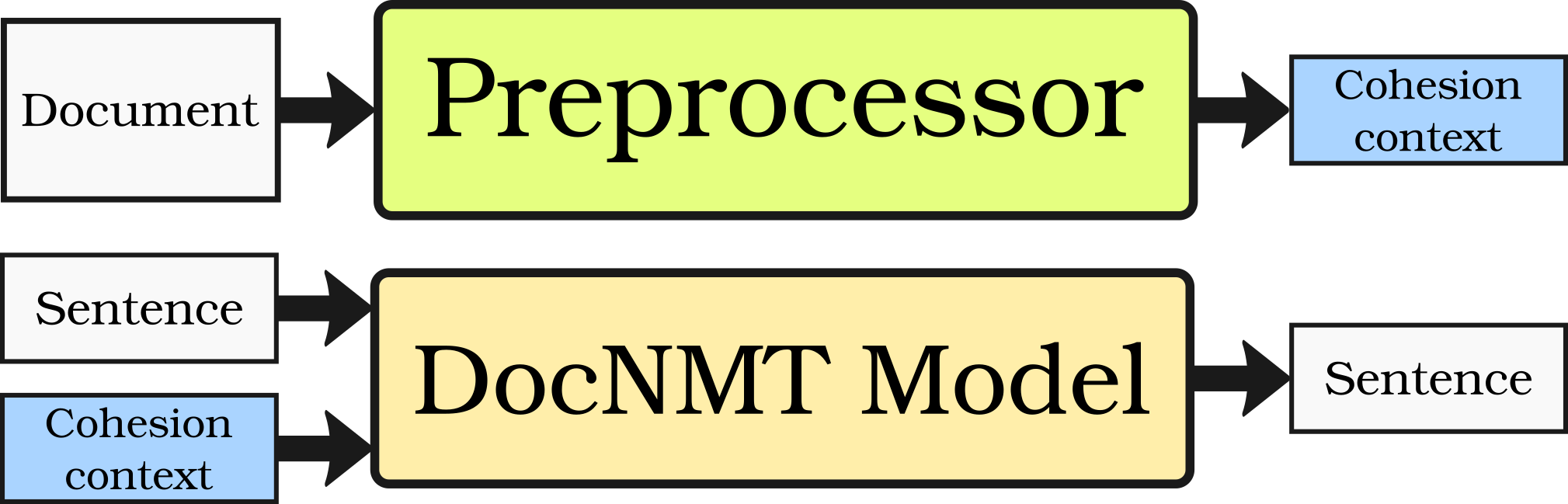}
\caption{A draft of our DocNMT pipeline architecture. We preprocess the document to mark cohesion features. Then we use the output as the data for our model.}
\label{fig:model}
\end{figure}

Next, we aim to build a DocNMT system which utilises marking of cohesion phenomena to make more cohesive translation choices\footnote{Including elliptical structures in this step will depend on the result of the first experiment.} (\autoref{fig:model}). We apply the insights from previous research, namely that the Transformer model may track cohesion phenomena when given enough context \citep{voita-etal-2018-context}, that context preprocessing stabilises performance of contextual MT models \citep{kim-etal-2019-document}, solutions to the problem of long inputs in DocNMT \citep[e.g.][]{ma-etal-2020-simple, Sun2020}, and finally our own analysis of the problem.

\subsection{Applying Evaluation Metrics to Cohesion and Speaker Phenomena}
Addressing RQ3 will involve testing the hypothesis that current common and SOTA automatic evaluation metrics fail to successfully reward translations which preserve contextual phenomena and, similarly, fail to punish those which do not.

We will develop a document-level test set of dialogue utterances in five languages, annotated for contextual phenomena. For each phenomenon, we will modify the reference translations to prepare several variations: one where all marked phenomena are translated correctly, another one where only 90\% is translated correctly, then 80\% etc. up to 0\%. We will prepare a set of common and SOTA MT evaluation metrics and use them to produce scores for all variants, for all phenomena. If there exists a metric which gives a consistently lower score the more a phenomenon is violated, for all phenomena, then our hypothesis is incorrect and we will use that metric for evaluation in experiments. Otherwise, we will develop our own metric.

The aforementioned test set will also be converted to a contrastive test suite \citep{muller-etal-2018-large} and submitted as an evaluation method to WMT News Translation task. The data to be used here is a combination of the Serial Speakers dataset \citep{Bost2020Serial} and exports from OpenSubtitles \citep{lison-tiedemann-2016-opensubtitles2016}, yielding 5.6k utterances total, split into scenes and parallel in five languages.

We hope that this work will substantiate the flaws of sentence-level evaluation and prompt the community to work on context-inclusive methods.

\section{Conclusions}
This work is the proposal of a PhD addressing PersNMT and DocNMT in the dialogue domain. We have presented evidence that sentence-level MT models make cohesion- and coherence-related errors and offered several approaches via which we aim to tackle this problem. We plan to conduct extensive experiments to analyse the problem of ellipsis translation and of the use of sentence-level evaluation metrics to evaluate contextual phenomena. The outcome of this work will also include publicly available test suites, a document-level translation model, a personalised translation model and a context-aware evaluation metric.

\section*{Acknowledgements}
This work was supported by the Centre for Doctoral Training in Speech and Language
Technologies (SLT) and their Applications funded by UK Research and Innovation [grant number EP/S023062/1].
\bibliography{references}
\bibliographystyle{acl_natbib}
\appendix
\section{Other examples}
In this section we present an extended set of examples supporting our hypotheses stated in the main proposal. All examples in \autoref{fig:more_ellipsis}, \autoref{fig:lex_coh} and \autoref{fig:more_reference} show examples of mistranslated sentences where the error was related to a specific phenomenon: ellipsis in \autoref{fig:more_ellipsis}, lexical cohesion in \autoref{fig:lex_coh} and reference in \autoref{fig:more_reference}. \autoref{fig:situation}, instead of highlighting translation errors, shows how a sentence in English can have several different translation candidates depending on the extra-textual context embedded in the situation (the corresponding translations are reference translations rather than MT-generated ones).

\begin{figure*}[t!]
\begin{tabular}{ll}
        \textbf{Context}    & What would they use it for?                                                                                                            \\
        \textbf{Antecedent} & [They would use it for]                                                                                                                \\
        \textbf{EN}         & Grabbing the balls of a spy.                                                                                                           \\
        \textbf{PL\textsubscript{MT}} & \textcolor{BrickRed}{Łapie} szpiega za jaja. \textit{(`He/she/they grab(s) the balls of a spy')}                                                                                                                 \\
        \textbf{PL\textsubscript{ref}}         & \textcolor{OliveGreen}{Żeby łapać} szpiega za jaja. \textit{(`For grabbing the balls of a spy')}                                                                        \\ \hline
        \textbf{Context}    & A big, dumb, balding, North American ape with no chin.                                                                                 \\
        \textbf{Antecedent} & [with]                                                                                                                                 \\
        \textbf{EN}         & And a short temper.                                                                                                                    \\
        \textbf{PL\textsubscript{MT}} & I \textcolor{BrickRed}{krótki temperament}\textsubscript{nominative}. \textit{(`And a short temper.')}                                                                                                                 \\
        \textbf{PL\textsubscript{ref}}         & I \textcolor{OliveGreen}{z krótkim temperamentem}\textsubscript{instrumental}. \textit{(`And with a short temper.')}                                                                                \\ \hline
        \textbf{Context}    & (...) with a record of zero wins and 48 \underline{defeats}...                                                                         \\
        \textbf{Antecedent} & [a record of zero wins and 48]                                                                                                           \\
        \textbf{EN}         & Oh, correction. Humiliating \underline{defeats}, all of them by knockout--                                                             \\
        \textbf{PL\textsubscript{MT}} & Oh, korekta. \textcolor{BrickRed}{Upokarzające porażki}, \textcolor{BrickRed}{wszystkie}\textsubscript{nominative} przez nokautowanie...             \\
        & \textit{(`Oh, correction. Humiliating defeats, all of them by knockout...')} \\
        \textbf{PL\textsubscript{ref}}         & Oh, korekta. \textcolor{OliveGreen}{Upokarzających porażek}, \textcolor{OliveGreen}{wszystkich}\textsubscript{genitive} przez nokautowanie... \\
        & \textit{(`Oh, correction. Humiliating defeats, all of them by knockout...')} \\ \hline
\textbf{Context}    & \enquote{I've only got two cupcakes for the three of you.}                                    \\
\textbf{Antecedent} & [two cupcakes]                                                                                \\
\textbf{EN}  &  \enquote{Just take mine [=my cupcake].}                                                              \\
\textbf{DE\textsubscript{MT}} & \enquote{Nimm einfach \textcolor{BrickRed}{meine} [=mine\textsubscript{fem}].}                 \\
\textbf{DE\textsubscript{ref}}  & \enquote{Nimm einfach \textcolor{OliveGreen}{meinen} [=mine\textsubscript{masc}].}                       \\
\end{tabular}
\caption{Examples of translations where resolving ellipsis is crucial to generating a correct translation hypothesis. \textbf{Context} is the utterance containing the antecedent, and \textbf{Antecedent} is the content which is elided in the current utterance. In the first two examples from the top, the Polish translation requires including part of the antecedent in order to maintain cohesion. In the third example from the top, the antecedent decides the inflection of all the words relating to the word \textit{defeats} which is repeated in the current utterance. Finally, the bottom example contains nominal ellipsis, and the model uses an incorrect inflection of \textit{mein} since it fails to make the connection with the antecedent.}
\label{fig:more_ellipsis}
\end{figure*}

\begin{figure*}[t!]
\begin{tabular}{ll}
\textbf{EN}         & \begin{tabular}[c]{@{}l@{}}\enquote{Sorry, Dad. I \underline{know you mean well}.} \enquote{Thanks for \underline{knowing I mean well}.}\end{tabular}               \\
\textbf{PL\textsubscript{MT}} & \begin{tabular}[c]{@{}l@{}}\enquote{Przepraszam tato. \textcolor{BrickRed}{Wiem}, że chcesz dobrze.} \enquote{Dzięki, że \textcolor{BrickRed}{wiedziałeś}, że chcę dobrze.} \end{tabular} \\
\textbf{PL\textsubscript{ref}}         & \begin{tabular}[c]{@{}l@{}}\enquote{Przepraszam tato. \textcolor{OliveGreen}{Wiem}, że chcesz dobrze.} \enquote{Dzięki, że \textcolor{OliveGreen}{wiesz}, że chcę dobrze.} \end{tabular}     \\ \hline
\textbf{EN}         & \begin{tabular}[c]{@{}l@{}}\enquote{You're a \underline{dimwit}.}\\ \enquote{Maybe so, but from now on... this \underline{dimwit} is on easy street.}\end{tabular}    \\
\textbf{PL\textsubscript{MT}} & \begin{tabular}[c]{@{}l@{}}\enquote{Jesteś \textcolor{BrickRed}{głupcem}.} \textit{(`You're a fool.')}\\ \enquote{Może i tak, ale od teraz ... ten \textcolor{BrickRed}{głupek} \textit{(dimwit)} jest na łatwej ulicy.}\end{tabular}   \\
\textbf{PL\textsubscript{ref}}         & \begin{tabular}[c]{@{}l@{}}\enquote{Jesteś \textcolor{OliveGreen}{głupkiem}.}\textit{(`You're a dimwit.')}\\ \enquote{Może i tak, ale od teraz ... ten \textcolor{OliveGreen}{głupek} \textit{(dimwit)} jest na łatwej ulicy.}\end{tabular} 
\end{tabular}
\caption{Examples of mistranslated lexical cohesion. In the top example, although the MT model managed to translate most of the repeated phrase in the same way, it failed to maintain the verb \textit{know} in the present tense. In the bottom example a different translation of \textit{dimwit} is used in the two utterances. Note that it is okay for a model to give a different hypothesis to a word than the human translator would, as long as it agrees with the source \textit{and} is cohesive with the rest of the text (i.e. all occurrences of the word are translated in the same way).}
\label{fig:lex_coh}
\end{figure*}

\begin{figure*}[t!]
\begin{tabular}{ll}
        \textbf{EN}         & The \underline{grabber}. What would they use \underline{it} for?                     \\
        \textbf{DE\textsubscript{MT}} & Der Grabber\textsubscript{masc}. Wofür würden sie \textcolor{BrickRed}{es}\textsubscript{neut} verwenden?                  \\
        \textbf{DE\textsubscript{ref}}         & Der Grabber\textsubscript{masc}. Wofür würden sie \textcolor{OliveGreen}{ihn}\textsubscript{masc} verwenden?                 \\ \hline
        \textbf{EN}         & Leave \underline{ideology} to the armchair generals. \underline{It} does me no good. \\
        \textbf{PL\textsubscript{MT}} & Ideologię\textsubscript{fem} zostawcie generałom foteli. Nic mi \textcolor{BrickRed}{to}\textsubscript{neut} nie da.      \\
        \textbf{PL\textsubscript{ref}}         & Ideologię\textsubscript{fem} zostawcie generałom foteli. Nic mi \textcolor{OliveGreen}{ona}\textsubscript{fem} nie da.    
\end{tabular}
\caption{Examples of mistranslated multi-sentence dialogue where reference is the violated phenomenon. In both examples, the gender of the referent is different in source and target languages, therefore the pronoun which refers to it is mistranslated.}
\label{fig:more_reference}
\end{figure*}

\begin{figure*}[t]
\begin{tabular}{p{0.2\linewidth}p{0.75\linewidth}}
\textbf{EN}                     & I never expected to be involved in every policy or decision, but I have been completely cut out of everything. \\
\textbf{PL (fem)}               & Nigdy nie oczekiwał\underline{am} wglądu w każdą decyzję, ale został\underline{am} odcięta od wszystkiego.                             \\
\textbf{PL (masc)}              & Nigdy nie oczekiwał\underline{em} wglądu w każdą decyzję, ale został\underline{em} odcięty od wszystkiego.                             \\ \hline
\textbf{EN}                     & And who have you called, by the way ?                                                                          \\
\textbf{PL (to masc)}           & Do kogo już dzwoni\underline{łeś}?                                                                                         \\
\textbf{PL (to fem)}            & Do kogo już dzwoni\underline{łaś}?                                                                                         \\
\textbf{PL (to Plural)}         & Do kogo już dzwoni\underline{liście}?                                                                                      \\
\textbf{PL (to Plural\textsubscript{fem})} & Do kogo już dzwoni\underline{łyście}?                                                                                      \\ \hline
\textbf{EN}                     & He was shot previous to your arrival?                                                                          \\
\textbf{PL (formal)}            & Został postrzelony przed \underline{pana} przyjazdem?                                                                      \\
\textbf{PL (informal)}          & Został postrzelony przed \underline{Twoim} przyjazdem?                                                                    
\end{tabular}
\caption{Examples of situation phenomena that can occur in text: speaker gender agreement (top), addressee gender agreement (middle), formality (bottom).}
\label{fig:situation}
\end{figure*}

\end{document}